\documentclass[journal]{IEEEtran}

\usepackage{graphicx}
\usepackage{amsmath,amssymb} 
\usepackage{url}

\usepackage{breakurl}
\usepackage[breaklinks]{hyperref}
\usepackage{array}
\usepackage{float}
\usepackage{booktabs}
\usepackage{comment}
\usepackage{subcaption}
\usepackage{color}
\usepackage{flushend}
\usepackage{epstopdf}
\usepackage{multirow}
\usepackage{bm}
\usepackage{paralist}
\usepackage{color}
\usepackage{hyperref}

\ifCLASSINFOpdf
\else
\fi

\begin{document}
	
%
\title{Multi-Scale Semantics-Guided Neural Networks for Efficient Skeleton-Based Human Action Recognition}
%
%
%

\author{{Pengfei Zhang},
	Cuiling Lan,~\IEEEmembership{Member,~IEEE,}
	Wenjun Zeng,~\IEEEmembership{Fellow,~IEEE,}
	Junliang Xing,~\IEEEmembership{Senior Member,~IEEE,}
	Jianru Xue,~\IEEEmembership{Member,~IEEE,}
	Nanning Zheng,~\IEEEmembership{Fellow,~IEEE}
	
\thanks{P. Zhang, J. Xue, and N. Zheng are with Xian Jiaotong University, Xian, Shannxi, P. R. China. E-mail: zpengfei@stu.xjtu.edu.cn, \{jrxue,nnzheng\}@mail.xjtu.edu.cn}
\thanks{C. Lan, and W. Zeng are with Microsoft Research Asia, Beijing, P. R. China. \protect E-mail: \{culan,wezeng\}@microsoft.com}
\thanks{J. Xing is with the National Laboratory of Pattern Recognition, Institute of Automatic, Chinese Academy of Sciences, Beijing, P. R. China. E-mail: jlxing@nlpr.ia.ac.cn}
\thanks{Corresponding authors: Jianru Xue}}

\maketitle

\begin{abstract}
Skeleton data is of low dimension. However, there is a trend of using very deep and complicated feedforward neural networks to model the skeleton sequence without considering the complexity in recent year. In this paper, a simple yet effective multi-scale semantics-guided neural network (MS-SGN) is proposed for skeleton-based action recognition. We explicitly introduce the high level semantics of joints (joint type and frame index) into the network to enhance the feature representation capability of joints. Moreover, a multi-scale strategy is proposed to be robust to the temporal scale variations. In addition, we exploit the relationship of joints hierarchically through two modules, {\it i.e.}, a joint-level module for modeling the correlations of joints in the same frame and a frame-level module for modeling the temporal dependencies of frames. With an order of magnitude smaller model size than most previous methods, MS-SGN achieves the state-of-the-art performance on the NTU60, NTU120, and SYSU datasets. 	
\end{abstract}

\IEEEpeerreviewmaketitle
\section{Introduction}

\begin{figure}[!t]
	\begin{center}
		\includegraphics[width=0.99\linewidth]{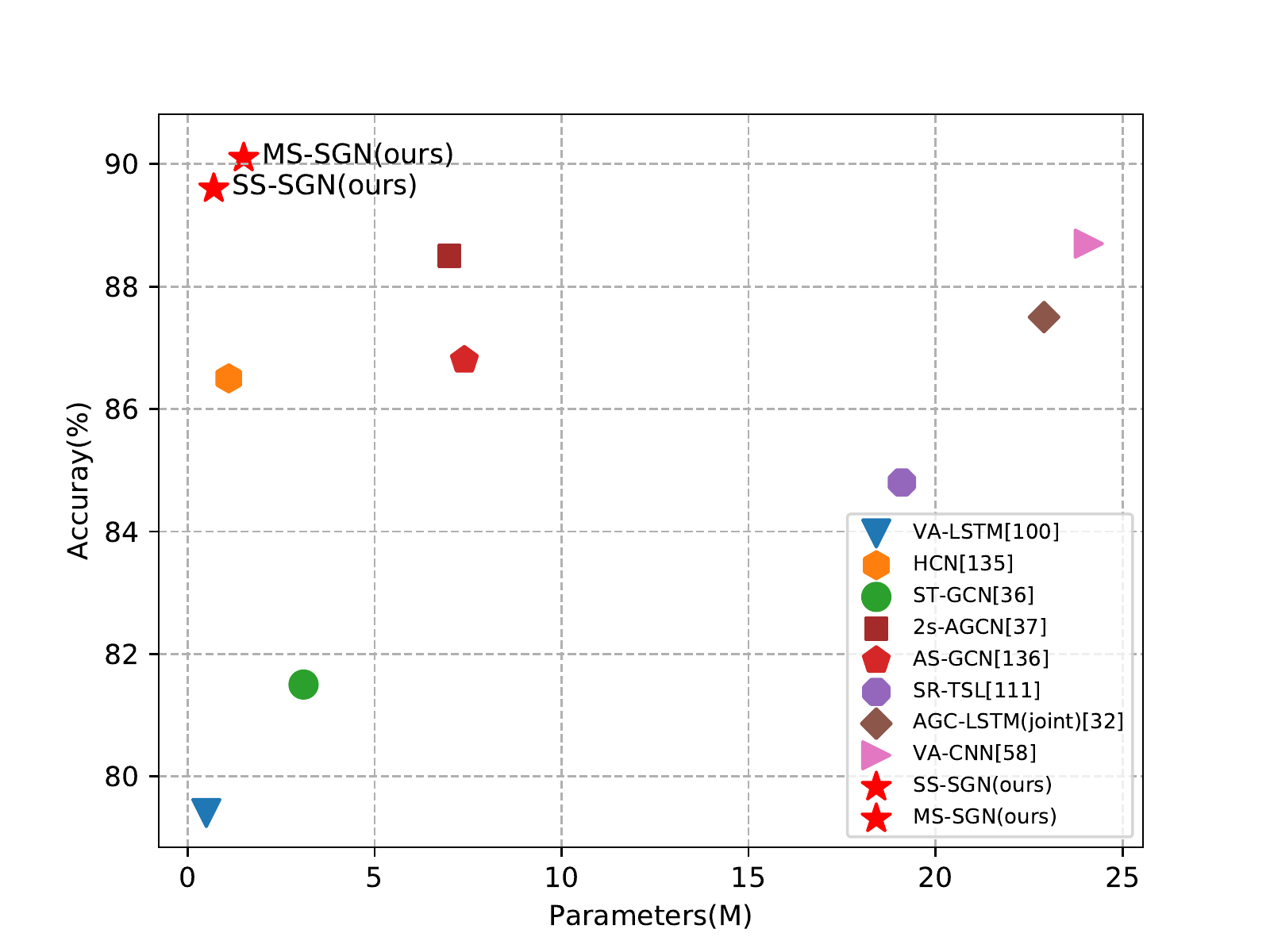}
	\end{center}
	\vspace{-4mm}
	\caption{Comparisons of different methods on NTU60 (the CS setting) in terms of accuracy and the number of parameters. The proposed SGN model achieves the best performance with an order of magnitude smaller model size.}
	\label{fig:paras}
\end{figure}

\IEEEPARstart{H}{uman} action recognition is a hot topic in computer vision, which has a lot of application scenarios, {\it e.g.}, human-computer interaction, video retrieval, and video understanding \cite{poppe2010survey, weinland2011survey, aggarwal2011human}. Skeleton-based action recognition \cite{yun2012two, du2015hierarchical, shahroudy2016ntu, zhang2019view} has attracted increasing interests in recent years. Skeleton is a type of well structured data with each person represented by a set of key joints, where each joint is identified by a joint type, a frame index, and a 3D position. Skeleton-based representation has several advantages/characteristics. First, skeleton is a high level representation, where the pose and motion of the human body is abstracted by some key joints. Biologically, human are able to distinguish different action categories by observing the motion and configuration of joints even without the help of appearance information \cite{johansson1973visual}. Second, skeleton data is easily available nowadays because of the cost effective depth cameras \cite{zhang2012microsoft} and pose estimation models \cite{shotton2011real,cao2017realtime,sun2019deep}. Third, compared with RGB video, the skeleton representation is robust to variation of appearance. Fourth, thanks to the low dimensional representation of the human body, the complexity and computation cost is friendly. Besides, RGB video data is complementary to skeleton 
data, where the combination of the two types of data can lead to better performance of action recognition \cite{song2018skeleton}. In this work, we focus on skeleton-based action recognition.

\begin{figure*}[!t]
	\begin{center}
		\includegraphics[width=1\linewidth]{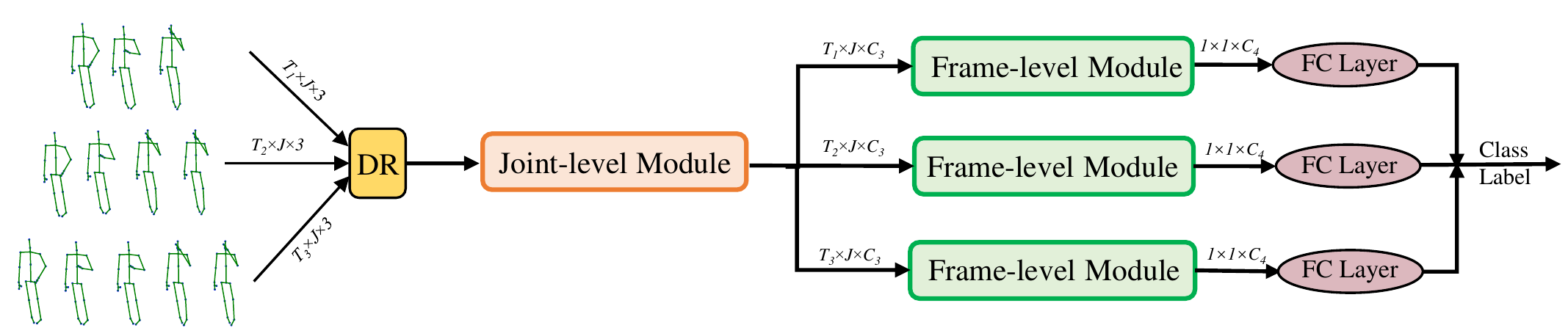}
	\end{center}
	\caption{Framework of the proposed end-to-end Multi-Scale Semantics-Guided Neural Network (MS-SGN). It consists of a dynamics representation (DR) module, a joint-level (JL) module and three frame-level (FL) modules. For the frame-level (FL) modules, we propose a multi-scale strategy by sampling each sequence in different temporal scales to form three  skeleton sequences. The DR module and JL module are shared for difference scales, while the FL modules are non-shared across scales.}
	\label{fig:framework}
\end{figure*}

\begin{figure}[!t]
	\begin{center}
		\includegraphics[width=0.99\linewidth]{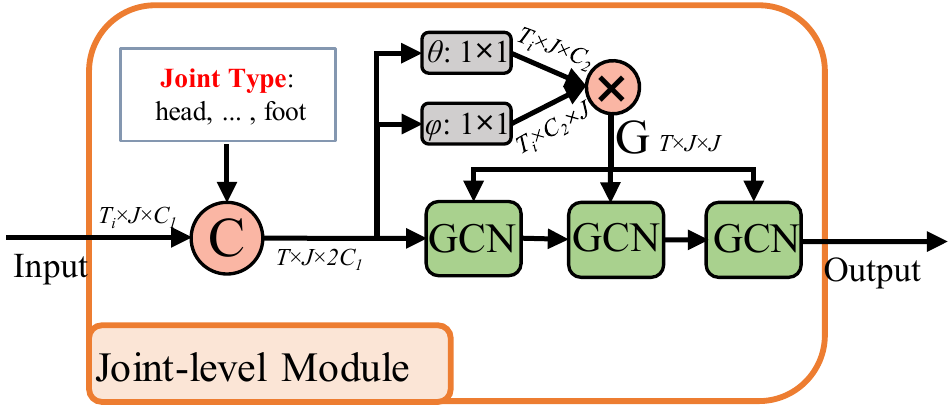}
	\end{center}
	\vspace{-4mm}
	\caption{Illustration of the joint-level module. The semantics of joint type are introduced in this module.}
	\label{fig:joint-level}
\end{figure}

\begin{figure}[!t]
	\begin{center}
		\includegraphics[width=0.99\linewidth]{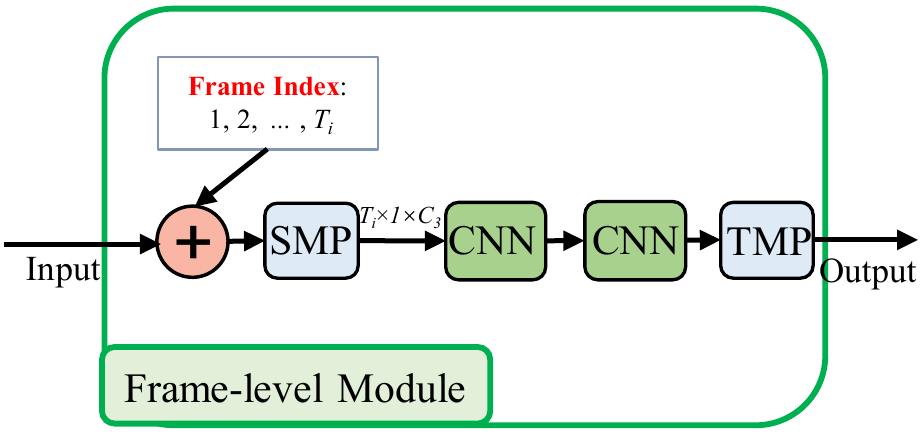}
	\end{center}
	\vspace{-4mm}
	\caption{Illustration of the frame-level module. The semantics of frame index is introduced in this module.}
	\label{fig:frame-level}
\end{figure}

In recent years, deep learning based models are designed to model the spatial and temporal dependencies of joints in a skeleton sequence for skeleton-based action recognition\cite{han2017space,wang2018rgb}. Various network structures have been exploited, such as Recurrent Neural Networks (RNN) \cite{du2015hierarchical, zhu2016co, shahroudy2016ntu, song2017end, zhang2017view, si2018skeleton}, Convolutional Neural Networks (CNN) \cite{ke2017new, zhang2019view, liu2017enhanced, weng2018deformable}, and Graph Convolutional Networks (GCN) \cite{yan2018spatial, si2018skeleton, tang2018deep}. In the early years, RNN/LSTM was widely used to exploit the short and long term temporal dynamics, in considering the powerful ability of RNNs for modeling temporal dependencies. Recently, there is a trend of using feedforward ({\it{i.e.}}, non-recurrent) convolutional neural networks for modeling sequences in speech, language \cite{oord2016wavenet,gehring2017convolutional,xiong2018microsoft,wang2018non}, and skeleton \cite{ke2017new,zhang2019view, liu2017enhanced, weng2018deformable} due to their superior performance. To make the CNN network suitable for skeleton-based action recognition, most approaches reorganize the skeleton sequence by mapping it to a skeleton map and resizing the skeleton map to a certain size ({\it e.g.} 224$\times$224) suitable for the input of a CNN network ({\it e.g.} ResNet50~\cite{he2016deep}), where the rows/columns of the skeleton map correspond to the different types of joints/frames indexes~\cite{ke2017new,zhang2019view, liu2017enhanced, weng2018deformable}. In these methods, long-term dependencies and semantic information are expected to be captured by the large receptive fields of deep networks. 
Recently, a lot of works tend to adopt GCN to exploit the spatio-temporal dependencies of the joints in the skeleton sequence by taking the coordinate of each joint as a node of a graph and conducting massage passing among the nodes with a pre-defined or content-dependent adjacency matrix \cite{yan2018spatial, shi2019two, si2019attention, tang2018deep, liu2020disentangling}. However, using only the coordinates of joints is less efficient to explore the structures of skeletons.   

Intuitively, the semantic information of skeleton sequence, {\it{i.e.}}, the joint type and the frame index, is beneficial for recognizing different action categories. The semantics together with dynamics ({\it{i.e.}}, 3D coordinates) reveal the spatial and temporal configuration/structure of human body joints. As we know, if the coordinates of two joints are the same but the semantics is different, they could carry/deliver very different information. For example, for a joint above the head, if this joint is hand, the action is likely to be \emph{raising hand}; if it is foot, the action may be \emph{kicking a leg}. Besides, the temporal order of frames is also discriminative to the recognition of actions. It is a key 
factor to distinguish some kinds of actions, {\it e.g.}, \emph{sitting down} and \emph{standing up}, where the only difference is the occurrence order of the frames. However, most approaches overlook the importance of the semantic information. In addition, different sequences present different temporal dynamics. Different subjects (with different ages, genders, and culture) tend to perform the same action in different motion patterns ({\it e.g.}, different speeds) and such diversity should be exploited. 

To address the above-mentioned problems, we propose a multi-scale semantics-guided neural network (MS-SGN) which explicitly exploits the semantics for efficient skeleton-based action recognition. The overall framework is shown in Fig.~\ref{fig:framework}. A hierarchical network is proposed, which explores the joint-level and frame-level dependencies of the skeleton sequence. It consists of a dynamics representation (DR) module, a joint-level (JL) module and three frame-level (FL) modules, where the DR module and JL module are shared by different temporal scale inputs, and the FL modules are non-shared. For the frame-level (FL) modules. 

As illustrated in Fig.~\ref{fig:joint-level}, for better joint-level correlation modeling, besides the dynamics, we incorporate the semantics of joint type ({\it e.g.}, `head', and `hip') to the GCN layers which enables the content adaptive graph construction and effective message passing among joints within each frame. As illustrated in Fig.~\ref{fig:frame-level}, for better frame-level correlation modeling, we incorporate the semantics of temporal frame index to the network. Particularly, we perform a Spatial MaxPooling (SMP) operation over all the features of the joints within the same frame to obtain frame-level feature representation. Combined with the embedded frame index information, two temporal convolutional neural network layers are used to learn feature representations for classification. 
In order to deal with the temporal variation of different subjects and actions, we feed the FL modules with multi-scale information at input-level. 
In addition, we build a strong baseline with high performance and efficiency by exploiting some technologies. We introduce fine-grained movement (see Fig. \ref{fig:finegrained} of joints in the same body part to the strong baseline, which enhances the ability of distinguishing the actions with similar poses. Thanks to the efficient exploration of semantic information, the hierarchical modeling, multi-scale information exploration, and the build of strong baseline, our proposed SGN achieves the state-of-the-art performance with a much smaller number of parameters. 

The main contributions are summarized as follows:

\begin{itemize}
    \setlength{\itemsep}{0pt}
    \item  We propose to explicitly explore the joint semantics (frame index and joint type) for efficient skeleton-based action recognition. Previous works overlook the importance of semantics and rely on deep networks with high complexity for action recognition. 
    \setlength{\itemsep}{0pt}
      \item We present a semantics-guided neural network (SGN) to exploit the spatial and temporal correlations at joint-level and frame-level hierarchically. 
	\item We explore the multi-scale information to be robust to the temporal variations of different subjects and actions.
	\item We develop a lightweight strong baseline, which is more powerful than most previous methods. We hope the strong baseline will be helpful for the study of skeleton-based action recognition. In the strong baseline, we explore the fine-grained movement of joints in the same body part, which significantly improves the ability of recognizing actions with similar skeletons.
    \setlength{\itemsep}{0pt}
\end{itemize}

With the above technical contributions, we have obtained a high performance skeleton-based action recognition model with high computational efficiency. Extensive ablation studies demonstrate the effectiveness of the proposed model designs. 

It should be mentioned that this paper is an extension
of our previous conference paper \cite{zhang2020semantics}. As an extension, we explore the multi-scale information to be robust to the temporal variations of different subjects and actions. Moreover, we explore the fine-grained movement of joints in the same body part, which significantly improves the ability of recognizing actions with similar skeletons. On the three largest benchmark datasets for skeleton-based action recognition, our proposed model consistently achieves superior performance while having an order of magnitude smaller model size when compared with many algorithms (see Fig.~\ref{fig:paras}).

\section{Related work}

Recently, skeleton-based action recognition is developing rapidly and attracting growing interests. Recent works using neural networks \cite{han2017space} have significantly outperformed traditional approaches that use hand-crafted features \cite{han2017space,xia2012view, wang2012mining, yu2014discriminative, garcia2017transition}. Therefore, we will limit our review to the deep learning based methods.

\noindent\textbf{Recurrent Neural Network based Methods.} Recurrent neural networks (RNNs), such as LSTM~\cite{hochreiter1997long} and GRU \cite{cho2014learning}, are often used to model the temporal dynamics of skeleton sequence through the recurrent connections in RNN. \cite{du2015hierarchical, shahroudy2016ntu, zhu2016co,song2017end, zhang2017view, zhang2018adding, zhang2019eleatt}. They tend to concatenate all joints (in a certain order) represented by 3D coordinates in a frame to form a vector, which will be taken as the input of different RNN structures without explicitly telling the networks which dimensions belong to which joint. To make the networks aware of the spatial structural information, some other RNN-based works design special structures in RNN. Shahroudy {\it et al.} divide the cell of LSTM  into five sub cells corresponding to five body parts, {\it i.e.}, torso, two arms, and two legs, respectively \cite{shahroudy2016ntu}. Liu {\it et al.} feed one type of joint at each step into their proposed spatial-temporal LSTM~\cite{liu2016spatio}. To some extent, they implicitly distinguish the different types of joints and body parts.    

\noindent\textbf{Convolutional Neural Network based Methods}. In recent years, convolutional neural networks are usually used to model the speech and language sequence, which lead to high performance and efficiency~\cite{oord2016wavenet,gehring2017convolutional,xiong2018microsoft,wang2018non, vaswani2017attention}. Compared with RNN-based methods, CNN-based methods show superiority in parallelism.  This holds for skeleton-based action recognition \cite{du2015skeleton, li2017skeleton, ke2017new,cao2018skeleton}, which transform the skeleton sequence to a skeleton map of some target size. CNNs, such as ResNet \cite{he2016deep}, will be used to explore the spatial and temporal dynamics of joints by taking the skeleton map as the input. Some works obtain the skeleton map by directly treating the joint coordinate ({\emph{x,y,z}}) as the R, G, and B channels of a pixel \cite{du2015skeleton, li2017skeleton}. Ke {\it et al.} transform the skeleton sequence to four 2D skeleton maps, which are represented by the relative position between four selected reference joints ({\it i.e.}, the left/right shoulder, the left/right hip) and other joints \cite{ke2017new}. Skeleton is well structured data with 
each joint owning its unique high level semantics, {\it i.e.}, frame index and joint type. However, the kernels/filters of CNNs are translation invariant \cite{long2015fully} and thus cannot directly perceive the semantics from such input skeleton maps. To be aware of such semantics, those CNN-based works tend to use a deep network with large receptive fields, which results in low computational efficiency.

\noindent\textbf{Graph Convolutional Network based Methods.} Graph convolutional networks \cite{kipf2016semi}, which have been proven to be effective for processing structured data, are suitable for modeling the dependencies of structural skeleton sequence. Yan {\it et al.} propose a spatial and temporal graph convolutional network, where each joint is treated as a node of the graph \cite{yan2018spatial}. The adjacency matrix is pre-defined by human based on prior knowledge. For example, if two joints are adjacent within the human body, the weight between them is high in the adjacency matrix. To enhance predefined graph and make it more suitable for action recognition, Tang {\it et al.} not only assign high weight value between physical connected joints, but also high weight value between certain physical disconnected joints. For example, a high weight value between two hands is vital for the action of \emph{writing}.  Liu {\it et al.} model the long-range dependencies of joints better by forcing the closer and further nodes to have the same influence on the current joints \cite{liu2020disentangling}. However, the pre-defined graph is not optimal for all actions. A SR-TSL model \cite{si2018skeleton} is proposed to learn the graph edge of five human body parts within each frame using a data-driven method instead of leveraging human definition. Shi {\it et al.} adopt non-local block to obtain graph for every skeleton sequence based on its content, followed by message passing in GCN layers with the learned graph \cite{shi2019two}.However, these data-driven methods overlook the importance of the informative semantics in learning the graph edge and message passing, which makes the networks less efficient.

\noindent\textbf{Explicit Exploration of Semantics Information.} In other fields,  {\it e.g.}, machine translation and image recognition,  some works have exploited the semantics explicitly and demonstrated its effectiveness \cite{vaswani2017attention, zheng2019learning}. Ashish {\it{et al.}} explicitly encode the position of the tokens in the sequence to make use of the order of the sequence in machine translation tasks \cite{vaswani2017attention}. Zheng {\it{et al.}} encode the group index into convolutional channel representation to preserve the information of group order \cite{zheng2019learning}. For skeleton-based action recognition, however, the important high level semantics of joint type and frame index is overlooked. In our work, we propose to explicitly encode the joint type and frame index to the representation of joint, which preserves the important spatial and temporal body structure and makes the networks more efficient. As an initial attempt to explore such semantics, we hope it will inspire more investigation and exploration in the community.

\section{Semantics-Guided Neural Networks}

A joint of the skeleton sequence can be identified by its semantics (joint type and frame index) and its dynamics (position/3D coordinates, velocity and fine-grained movement). Semantics is vital to preserve the important information of spatial and temporal structure and configuration of a human body. 
Previous works \cite{ke2017new, du2015skeleton, zhang2019view,yan2018spatial,si2018skeleton,shi2019two}, however, typically overlook the semantics. Moreover, the scalable motion dynamics are in general under-explored in skeleton based action recognition. For examples, for the same action, the pattern/speed of acting by a young man could be rather different from that of an old man. 

We propose a multi-scale semantics-guided neural network (MS-SGN) for skeleton-based action recognition and show the overall end-to-end framework in Fig.~\ref{fig:framework}. For a skeleton sequence, we first identify a joint in dynamics representation (DR) module by combining position, velocity, and fine-grained movement information together. Then, a joint-level (JL) module exploits the correlations of joints in the same frame under the guidance of the semantics of joint type. Afterward, three frame-level (FL) modules exploit the correlations of frames by exploring the temporal correlations of different scales under the guidance of the semantics of temporal index. For different scale information, the DR module and JL module are shared while the FL modules are non-shared across scales. We describe the details of the framework in the following subsections.

We denote a skeleton sequence as a set of joints $\mathcal{S}$ = \{$X_t^k$ $\mid$ $t = 1, 2, \dots, T; k= 1, 2, \dots, J$\}, where $X_t^k$ denotes the joint of type $k$ at time $t$. $T$ denotes the number of frames of the skeleton sequence and $J$ denotes the total number of joints of a human body in a frame. It is noted that $T$ differs for different scale information and we omit the denotation of scale for simplicity.

\subsection{Dynamics Representation Module}

\begin{figure}[!t]
	\begin{center}
		\includegraphics[width=0.8\linewidth]{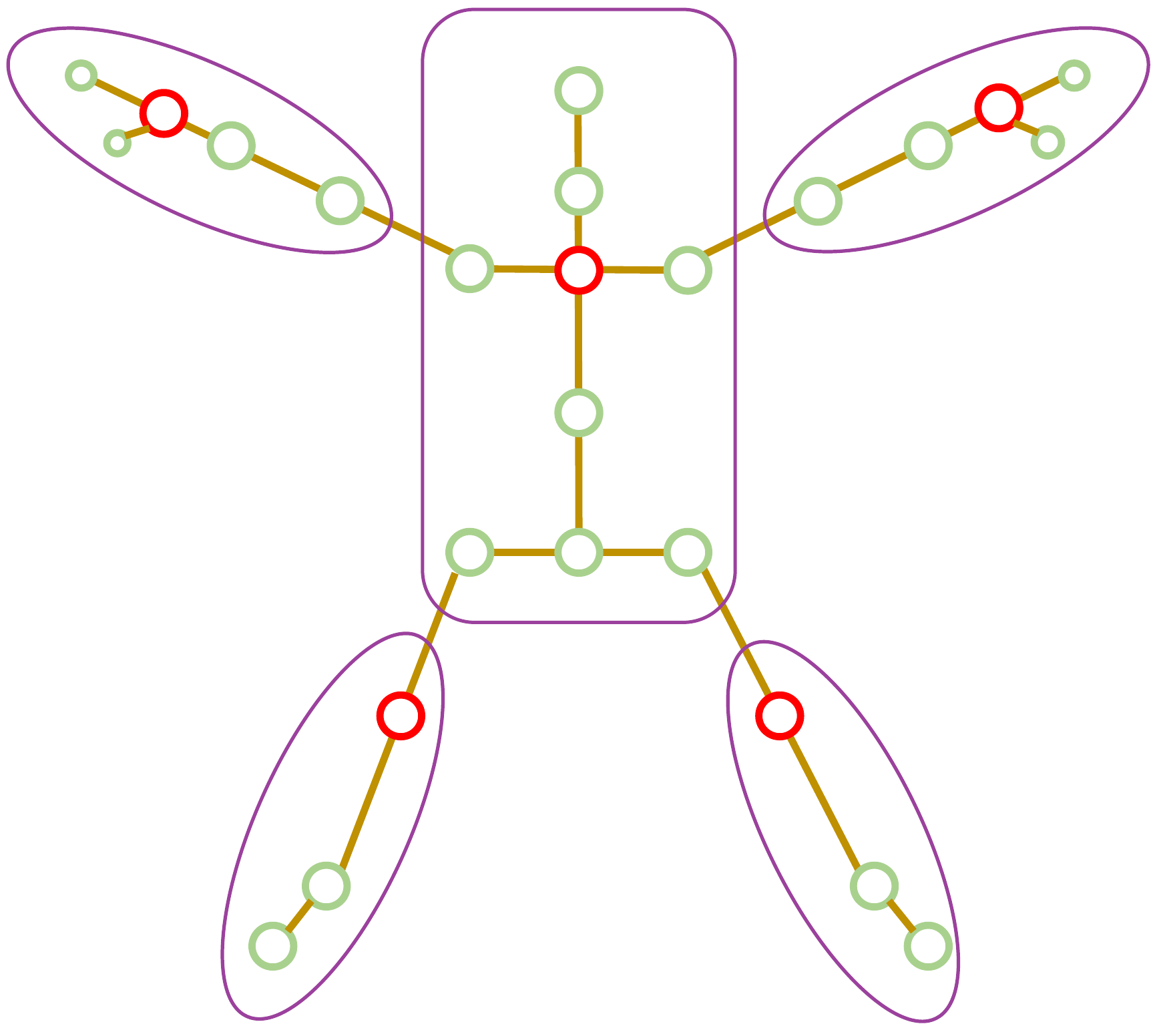}
	\end{center}
	\vspace{-4mm}
	\caption{Illustration of the selected reference joints (marked by red) for each body part (marked by purple) for fine-grained movement modeling. }
	\label{fig:finegrained}
\end{figure}

For a given joint $X_t^k$, we define its dynamics by the position $\mathbf{p}_{t,k} = (x_{t,k}, y_{t,k}, z_{t,k})^T \in \mathbb{R}^3$ in the 3D coordinate system, the velocity $\mathbf{v}_{t,k} = \mathbf{p}_{t,k} - \mathbf{p}_{t-1,k}$, and the fine-grained movement $\mathbf{m}_{t,k} = \mathbf{p}_{t,k} - \mathbf{r}_{t,n}$ and $\mathbf{m}^{'}_{t,k} = \mathbf{v}_{t,k} - \mathbf{r}^{'}_{t,n}$, where ${r}_{t,n}$ and ${r}^{'}_{t,n}$ denote the referent joint of $n^{th}$ body part in time $t$ for position and velocity, respectively. As illustrated in Fig.~\ref{fig:finegrained}, we divide the whole human body into 5 body parts. For each body part, we select one joint as its referent joint, {\it i.e.}, left hand, right hand, left knee, right knee, and spine. The fine-grained movements capture the fine-gained relations and motions. We encode/embed the position, velocity, and the fine-grained movement to features, {\it i.e.}, $\widetilde{\mathbf{p}_{t,k}}$, $\widetilde{\mathbf{v}_{t,k}}$, $\widetilde{\mathbf{m}_{t,k}}$, and $\widetilde{\mathbf{m}^{'}_{t,k}}$ respectively, in the same high dimensional space. We fuse them together by summation as
\begin{equation}
\textbf{z}_{t,k} = \widetilde{\mathbf{p}_{t,k}}  +  \widetilde{\mathbf{v}_{t,k}} + \widetilde{\mathbf{m}_{t,k}} + \widetilde{\mathbf{m}^{'}_{t,k}} \in \mathbb{R}^{C_{1}},
\end{equation}
\label{eq:fuse}
where $C_{1}$ denotes the number of dimensions of a joint representation. We use the same network structure for the feature encoding of the four kinds of dynamical information, respectively. Taking the embedding of position as an example, two fully connected layers (FC) are used to encode the position $\mathbf{p}_{t,k}$ to high dimension space, which is formulated as follows:
\begin{equation}
    \widetilde{\mathbf{p}_{t,k}} = \sigma(W_2(\sigma(W_1\mathbf{p}_{t,k} + \mathbf{b}_1))+ \mathbf{b}_2),
    \label{equ:mapping}
\end{equation}
where $W_1 \in \mathbb{R}^{C_1 \times 3}$ and $W_2 \in \mathbb{R}^{C_1 \times C_1}$ are weight matrices, $\mathbf{b}_1$ and $\mathbf{b}_2$ are the bias vectors, $\sigma$ denotes the ReLU activation function \cite{nair2010rectified}. Similarly, we obtain the embedding for velocity as $\widetilde{\mathbf{v}_{t,k}}$ and $\widetilde{\mathbf{m}_{t,k}}$.

 

\subsection{Joint-level Module}

The joint-module is designed for exploiting the dependencies of joints in the same frame, which is illustrated in Fig.~\ref{fig:joint-level}. We adopt graph convolutional networks (GCN) to explore the correlations for the structural skeleton data. One key to apply GCN to skeleton-based action recognition is how to design suitable graph. Some previous GCN-based approaches obtain the graph by taking the joints as nodes with a pre-defined adjacency matrix based on prior knowledge \cite{yan2018spatial}  or a learned adaptive graph based on the input content \cite{shi2019two}. We adopt the content adaptive strategy to obtain the adjacency matrix. Differently, we incorporate the semantics of joint type to the GCN layers for more effective learning.

We enhance the power of GCN layers by making full use of the semantics from two aspects. First, the semantics of joint type and the dynamics are combined together to learn the graph connections among the nodes (different joints) within a frame. The joint type information is helpful for learning suitable adjacent matrix ({\it i.e.}, relations between joints in terms of connecting weights). Take two source joints, \emph{foot} and \emph{hand}, and a target joint \emph{head} as an example, intuitively, the connection weight value from \emph{foot} to \emph{head} should be different from the value from \emph{hand} to \emph{head} even when the dynamics of \emph{foot} and \emph{hand} are the same. Second, as part of the information of a joint, the semantics of joint types takes part in the message passing process in GCN layers. Previous GCN is position-independent and it has no idea of the structural information of skeleton, which is less efficient for recognizing actions. In contrast, we take advantage of the semantics of joint type to make the GCN know the structural information of skeleton during massage passing.

We denote the type of the $k^{th}$ joint (also referred to as type $k$) as a one-hot vector $\mathbf{j}_{k} \in \mathbb{R}^{d_j}$, where the $k^{th}$ dimension is one and the others are all zeros. Similar to the encoding of position as in (\ref{equ:mapping}), we obtain the embedding of the $k^{th}$ joint type in high dimension space as $\widetilde{\mathbf{j}_{k}} \in \mathbb{R}^{C_1}$. 

For each skeleton frame consisting of $J$ joints, we build a graph of $J$ nodes. To enhance the representation ability of a joint, we encode it by fusing its dynamics and semantics of the joint type together, {\it i.e.}, $\textbf{z}_{t,k}$ = [$\textbf{z}_{t,k}, \widetilde{\mathbf{j}_{k}}] \in \mathbb{R}^{2C_{1}}$. All the joints of frame $t$ are then represented by $Z_{t} = (\textbf{z}_{t,1}; \cdots; \textbf{z}_{t,J}) \in \mathbb{R}^{J \times 2C_1}$.

Similar to \cite{wang2018videos,wang2018non,shi2019two}, the edge weight from the $i^{th}$ joint to the $j^{th}$ joint in the same frame $t$ is computed by their similarity/affinity in the embeded space as
\begin{eqnarray}
      \!& S_t(i,j)=  {\theta}(\mathbf{z}_{t,i})^T {\phi}(\mathbf{z}_{t,j}),
    \label{equ:G}
\end{eqnarray}
where $\theta$ and $\phi$ are implemented by two FC layers, {\it i.e.}, ${\theta}(\mathbf{x}) = W_3 \mathbf{x} + \mathbf{b}_3 \in \mathbb{R}^{C_2} $ and $ ~{\phi}(\mathbf{x}) = W_4 \mathbf{x} + \mathbf{b}_4 \in \mathbb{R}^{C_2}$. 

By computing the affinities of all the joint pairs in the same frame based on (\ref{equ:G}), we obtain the adjacency matrix $S_t \in \mathbb{J\times J}$. Normalization using SoftMax as \cite{vaswani2017attention, wang2018non} is performed on each row of $S_t$ so that the sum of all the edge values connected to same target node is 1. The normalized adjacency matrix is denoted as $G_t$.
A graph convolution layer with the massage passing among nodes is formulated as
\begin{eqnarray}
\label{equ:P}
\begin{aligned}
            \!& Y_t =  G_{t}Z_{t}W_y, \\
            \!& Z'_t = Y_t+ Z_tW_z,
\end{aligned}
\end{eqnarray}
where $W_y$ and $W_z$ are transformation matrices. The weight matrices are shared for different temporal frames. $Z'_t$ is the output. Note that one can stack multiple residual graph convolution layers to enable further message passing among nodes with the same adjacency matrix $G_t$.

\subsection{Frame-level Module}

The frame-level module is designed for exploiting the dependencies across frames, which is shown in \ref{fig:frame-level}. To make the network aware of the order of frames, we incorporate the semantics of frame index to enhance the representation capability of a frame. 

Similar to joint type, we denote the frame index by a one-hot vector $\mathbf{f}_{t} \in \mathbb{R}^{d_{f}}$. With the similar way of encoding position to high dimension space as shown in (\ref{equ:mapping}), we obtain the embedding of the frame index as $\widetilde{\mathbf{f}_{t}} \in \mathbb{R}^{C_{3}}$. We denote the joint representation corresponding to joint type $k$ at frame $t$ with both the semantics of frame index and the learned feature as $\textbf{z}'_{t,k}$ = $\textbf{z}'_{t,k} + \widetilde{\mathbf{f}_{t}} \in \mathbb{R}^{C_{3}}$, where $\textbf{z}'_{t,k} = Z'_t(k,:)$.

A spatial MaxPooling layer is applied to merge the information of all joints in a frame. Therefore, joints in the same frame will be taken as a whole. The dimension of feature of the sequence is thus $T\times 1 \times C_{3}$. 
Two CNN layers are applied. The first CNN layer is a temporal convolution layer to model the dependencies of frames. The second CNN layer is used to enhance the representation capability of learned features by mapping it to a high dimension space with temporal kernel size of 1. After the two CNN layers, we apply a temporal MaxPooling layer to aggregate the information of all frames and obtain the sequence level feature representation of $C_{4}$ dimensions. This is then followed by a fully connected layer with Softmax to perform the classification.  

\subsection{Mulit-scale SGN}

To better explore the temporal dynamics/variations, we propose a multi-scale strategy. As we know, different subjects tend to perform actions in different speeds/patterns, which leads to the temporal variation of one action varies in a large range. In order to alleviate the influence of temporal variation, we sample each sequence in multi temporal scales and feed them to the proposed MS-SGN at input-level. As illustrated in Fig.~\ref{fig:framework}, in MS-SGN, we have multiple frame-level (FL) modules for tackling different temporal scaled actions. Note that DR module and JL module are shared for different scaled sequence while the FL modules not shared.

\section{Experiments}

In the following, we demonstrate the effectiveness of the proposed multi-scale semantics-guided neural networks for skeleton-based action recognition. We first describe the datasets and the implementation details in Subsection \ref{dataset} and \ref{implement}. In Subsection \ref{ablation}, we perform ablation studies to analyze how our model works. In Subsection \ref{compare}, we compare our SGN with the stat-of-the-art approaches on three benchmark datasets.

\subsection{Datasets}
\label{dataset}

\noindent\textbf{NTU60 RGB+D Dataset (NTU60) \cite{shahroudy2016ntu}}. This dataset is collected by the Kinect V2 camera for 3D action recognition. It contains 56,880 skeleton sequences in total, including 60 different action classes performed by 40 different subjects with different gender and age. There are 25 key joints in a human skeleton body and each joint is represented 3D coordinates. We use the same protocol settings \cite{shahroudy2016ntu}, {\it i.e.}, Cross Subject (CS) and Cross View (CV). For the CS setting, half of the 40 subjects are used for training and the rest for testing. For CV setting, the sequences captured by two of the three cameras are used for training and those captured by the other camera are used for testing. During training, 10\% of the training sequences are randomly selected as validation sequences for both the CS and CV settings as \cite{shahroudy2016ntu}.

\noindent\textbf{NTU120 RGB+D Dataset (NTU120) \cite{liu2019ntu}}. As an extension of NTU60 dataset, NTU120 dataset contains 114,480 skeleton sequences in total, including 120 action classes performed by 106 distinct human subjects. We use the same protocol settings \cite{shahroudy2016ntu}, {\it i.e.}, Cross Subject (C-Subject) and Cross Setup (C-Setup). For C-Subject setting, half of the 106 subjects are used for training and the rest for testing. For C-Setup setting, half of the setups are used for training and the rest for testing.

\noindent\textbf{SYSU 3D Human-Object Interaction Dataset (SYSU) \cite{hu2015jointly}}. It contains 480 skeleton sequences of 12 actions performed by 40 different subjects. There are 20 key joints in a human skeleton body and each joint is represented 3D coordinates. We use the same evaluation protocols as \cite{hu2015jointly}, {\it i.e.}, Cross Subject (CS) and Same Subject (SS) setting. For the CS setting, half of the subjects are used for training and the rest for testing. For the SS setting, half of the samples of each activity are used for training and the rest for testing.30-fold cross-validation are used during training and we show the mean accuracy for each setting \cite{hu2015jointly}.

\subsection{Implementation Details}
\label{implement}
\noindent\textbf{Network Setting}. In the dynamic representation (DR) module, the number of neurons of each FC layer is set to 64 ({\it i.e.}, $C_1=64$). Note that the weights of FC layers are non-shared for position, velocity, and fine-grained movement. To encode the joint type, the number of neurons of the two FC layers are both set to 64. To encode the frame index, the numbers of neurons of the two FC layers are set to 64 and 256, respectively and $C_3=256$. To compute the similarity in embed space as shown in (\ref{equ:G}), the number of neurons of each FC layer is set to 256, {\it i.e.}, $C_2=256$. In the joint-level module, the numbers of neurons of the three GCN layers are set to 128, 256, and 256, respectively. In the fame-level module, the number of neurons of the first CNN layer is set to 256 with kernel size of 3 along the temporal dimension, and the number of neurons of the second CNN layer is set to 512 with kernel size of 1 ({\it i.e.}, $C_{4}=512$). After each GCN or CNN layer, we apply batch normalization \cite{ioffe2015batch} followed by  ReLU nonlinear activation layer.

\noindent\textbf{Training}. All experiments are conducted on the Pytorch platform with one GPU card. We use the Adam \cite{kingma2014adam} optimizer with the initial learning rate of 0.001. The learning rate decays by a factor of 10 at the 60$^{th}$ epoch, the 90$^{th}$ epoch, and the 110$^{th}$ epoch, respectively. The training is finished at the 120$^{th}$ epoch. We use a weight decay of 0.0001. The batch sizes for NTU60, NTU120, and SYSU datasets are set to 64, 64 and 16, respectively. Label smoothing \cite{he2019bag} is utilized for all experiments and we set the smoothing factor to 0.1. Cross entropy loss for classification is used to train the networks.

\noindent\textbf{Data Processing}. Similar to \cite{zhang2017view}, sequence level translation based on the first frame is performed to be invariant to the initial positions. If one frame contains two persons, we split the frame into two frames by making each frame contain one human skeleton. 
During training, we segment the each skeleton sequence into 15, 20, and 25 clips equally to obtain multi-scale skeleton sequences, and randomly select one frame from each clip, which make the final skeleton sequences contains 15, 20, and 25 frames, respectively. 
During testing, similar to \cite{baradel2018glimpse}, we randomly create 5 new sequences for each single-scale skeleton sequence in the same manner and then predict the final action class using the mean score.

To alleviate the influence of view variation, we perform data argumentation in the training procedure, {{\it i.e.},} rotating the 3D skeletons randomly by some degrees at sequence level. For the NTU60 (CS setting), NTU120, and SYSU datasets, we randomly select three degrees (around $X$, $Y$, $Z$ axes, respectively) between [$-17^\circ, 17^\circ$] for a sequence. Considering the large view variation for NTU60 (CV setting), we randomly select three degrees between [$-30^\circ , 30^\circ$].

\subsection{Ablation Study}
\label{ablation}

In Subsection \ref{semantics}, we evaluate the effectiveness of explicitly exploiting semantics. The analysis of effectiveness of hierarchical model is presented in Subsection \ref{hierarchical}. The strong baseline and its analyses are introduced in Subsection \ref{strong}. We evaluate the effectiveness of fine-grained movement in Subsection \ref{sec:fine}. The analyses of the proposed multi-scale strategy is presented in \ref{sec:fine}. The visualization of the responses of the spatial MaxPooling layer is shown in Subsection \ref{vis}. We compare the number of parameters between the proposed MS-SGN and eight other state-of-the-art methods in Subsection \ref{efficiency}

Considering NTU120 is the largest dataset for skeleton-based action recognition, we perform our ablation study based on it. We design two modules, {\it i.e.}, single-scale semantics-guided neural network (SS-SGN), where each skeleton sequence contains 20 frames, and mulit-scale semantics-guided neural network (MS-SGN), where each skeleton sequence is sampled to have three temporal scales of 15, 20, and 25 frames, respectively. Both modules contain the fine-grained movement.

\subsubsection{Effectiveness of Exploiting Semantics}
\label{semantics}

Semantics contains the important structural information of a skeleton sequence which is important for skeleton-based action recognition. We use the SS-SGN to demonstrate the effectiveness of exploiting semantics by building eight neural networks and performing various experiments on the NTU120 dataset. Table \ref{tab:sem} shows the results. For the eight models, \emph{JT} denotes the semantics of joint type, \emph{FI} denotes the semantics of frame index, \emph{G} denotes the learning of graph (adjacency matrix), \emph{P} denotes the graph convolutional operations which enable the massage passing. \emph{T-Conv} denotes the temporal convolutional layer, {\it i.e.}, the first CNN layer of the frame-level module. Three GCN layers and two CNN layers are used in the joint-level (\textbf{JL}) module and the frame-level (\textbf{FL}) module, respectively. \emph{w} and \emph{w/o}  denote ``with" and ``without", respectively.

\noindent\textbf{Effectiveness of Exploiting Joint Type.}
To validate the effectiveness of the joint type on the joint-level module,  we design four models based on SS-SGN as shown in rows 1 to 4 in Table \ref{tab:sem}, where the semantics of temporal index is not used for all the four experiments. We explain the setting of one model in detail here, and the settings of the other three models can be obtained in a similar way. For the model ``JL(G w/o JT \& P w/o JT) \& FL", the semantics of joint type is not used for learning adjacency matrix ($G$) ({\it i.e.}, G w/o JT) and not used  for massage passing ($P$) in GCN layers ({\it i.e.}, P w/o JT). We have three main observations about the effectiveness of the semantics of joint type on the joint-level module as follows.


\noindent\textbf{1)} For the learning of adjacency matrix of the graph, with the help of the semantics of joint type, ``JL(\textbf{G w JT} \& P w/o JT) \& FL" outperforms ``JL(G w/o JT \& P w/o JT) \& FL" by 0.4 for the CS setting. Intuitively, if the types of the joints are not introduced into the model, it cannot distinguish the joints with the same coordinates even though their semantics are different, which results in inferior edge weight between those joints. The semantics of joint type is beneficial for learning graph edges.

\noindent\textbf{2)} For the message passing in GCN layers, with the help of the semantics of joint type, ``JL(G w/o JT \& \textbf{P w JT}) \& FL" is superior to ``JL(G w/o JT \& P w/o JT) \& FL" by 0.8\%. The reason is that GCN itself is not aware of the order (type) of joints which makes it hard to learn features of the skeleton data with high structural information. For example, the information contributed from foot joint and wrist joint to a target joint should be different even when the 3D coordinates of the two joints are the same during the message passing. Introducing the joint type information makes GCN more efficient. 

\noindent\textbf{3)}  Using the semantics of joint type for both learning adjacency matrix of the graph and the message passing at the same time (``JL(G w JT \& P w JT) \& FL") does not bring further benefits in comparison with ``JL(G w/o JT \& \textbf{P w JT}) \& FL". For message passing $Y_t=G_tZ_tW$ in (\ref{equ:P}), the gradient back-propagated to $G_t$ will also be influenced by $Z_t$ which contains joint type information. Actually, $G_t$ is aware of the joint type information implicitly even though we do not include joint type information in the similarity/affinity learning.

\begin{table}[t]
  \small
  \centering
  \caption{Effectiveness of exploiting semantics in the joint-level module (JL) and frame-level module (FL) on the CS setting of the NTU120 dataset in terms of accuracy (\%). All experiments are based on SS-SGN model.}
    \begin{tabular}{m{0.25\textwidth} p{0.075\textwidth}<{\centering}  p{0.075\textwidth}<{\centering}}
    \toprule
    Method  & \#Params(M) & CS \\
    \midrule
    JL(G w/o JT \& P w/o JT) \& FL & 0.66  &81.0 \\
    JL(G w JT \& P w/o JT) \& FL & 0.70    &81.4    \\
    JL(G w/o JT \& P w JT) \& FL & 0.68    &81.8  \\
    JL(G w JT \& P w JT) \& FL & 0.71      &\textbf{81.8}  \\
    \midrule
    JL \& FL(w/o T-Conv) w/o FI &0.58   & 79.8  \\
    JL \& FL(w/o T-Conv) w FI   &0.60    &\textbf{80.9} \\
    \midrule
    JL \& FL(w T-Conv) w/o FI & 0.71   &81.8  \\
    JL \& FL(w T-Conv) w FI & 0.73     &\textbf{82.3}  \\
    \bottomrule
    \end{tabular}
  \label{tab:sem}
\end{table}%


\noindent\textbf{Effectiveness of Exploiting Frame Index.} To study the influence of the frame index, we design two models based on the SS-SGN as shown in rows 5 and 6 in Table \ref{tab:sem}), where the temporal convolution is degraded by setting its kernel size to 1. ``JL \& FL(w/o T-Conv) w FI" denotes the model using the semantics of frame index. Both models have incorporated the semantics of joint type. 

Moreover, we investigate two models (rows 7 and 8 in Table \ref{tab:sem}) to study the influence of the frame index when the temporal convolution with kernel size of 3 is used. ``JL \& FL (w T-Conv) w FI" denotes the model using the semantics of frame index. Both models have incorporated the semantics of joint type.

For the effectiveness of the semantics of temporal index on the frame-level module, there are two observations.

\noindent\textbf{1)} When the temporal convolution is disabled ({\it i.e.}, filter kernel size is 1 instead of 3), ``JL \& FL(w/o T-Conv) w \textbf{FI}" outperforms ``JL \& FL(w/o T-Conv) w/o FI" by 1.1\% the CS setting. The frame index information ``tells" the network the frame order of skeleton sequence which is beneficial for action recognition. 

\noindent\textbf{2)} The frame index is helpful for temporal convolution. ``JL \& FL (w T-Conv) w \textbf{FI}" is superior to ``JL \& FL (w T-Conv) w/o FI" by 0.5\% for the CS setting. The benefits from the semantics of frame index are smaller than those models without temporal convoluitonal (with filter kernel size of 1). The main reason is the temporal convolutional layer enables the network to know the frame order of skeleton sequence to some extent through large kernel size. However, ``telling" the networks the semantics of frame index explicitly further improves the performance with negligible cost. We take the scheme ``JL \& FL (w T-Conv) w \textbf{FI}" as our final scheme, which is also referred to as ``SGN". 

In summary, the explicit modeling of the joint type information benefits the learning of adjacent matrices and the message passing in the GCN layers. The frame index information enables the model to efficiently exploit the information of sequence order.


\subsubsection{Effectiveness of Hierarchical Model}
\label{hierarchical}

\begin{table}[t]
\small
\centering
\caption{Effectiveness of our hierarchical model on the CS setting of the NTU120 dataset in terms of accuracy (\%).}
\begin{tabular}{m{0.22\textwidth} p{0.09\textwidth}<{\centering}  p{0.09\textwidth}<{\centering}}
\toprule
Method         & \#Params(M) & CS       \\
\midrule
SS-SGN w G-GCN(1x3)      & 0.72     & 81.4  \\
SS-SGN w G-GCN(3x3)      & 1.11     & 81.2  \\
SS-SGN      & 0.73    & \textbf{82.3}  \\
\bottomrule
\end{tabular}
\label{tab:model}
\end{table}

We hierarchically model the correlations of the joints by building the dependencies of joints in the same frame through joint-level module and building the dependencies of frames in the frame-level module, where joints in the same frame are taken as a whole.  To demonstrate the effectiveness of hierarchical design, we compare the proposed SS-SGN with two different non-hierarchical models and show the results in Table \ref{tab:model}.

``SS-SGN w G-GCN(1x3)" denotes a non-hierarchical scheme where we remove the spatial MaxPooling layer (SMP), and use the combined semantics ({\it i.e.}, joint type and frame index) and dynamics (position and velocity) in the GCN layers. Instead of constructing a graph for each frame, we build a global adaptive graph with all the joints in all the frames and conduct message passing among all those joints. In the frame-level module, the kernel size of the first CNN layer is set to 1x3, which is same with ``SS-SGN", which only models the dependencies of the joints temporally.  ``SS-SGN w G-GCN(3x3)" is similar to ``SS-SGN w G-GCN(1x3)" and the only difference is that the kernel size of the first CNN layers is set to 3x3 in the frame-level module, which models the dependencies of the joints spatially and temporally. Without the first CNN layer with large kernel to build temporal relations in the frame-level module, the performance of the above two non-hierarchical models will decrease dramatically. 


From Table \ref{tab:model}, we have the following observation.

Hierarchical design is much more powerful than non-hierarchical design in exploiting the dependencies of joints. Modeling the correlations of joints of the same frame by GCN is much more effective than modeling the correlations of all joints of all the frames. ``SS-SGN" is superior to ``SS-SGN w G-GCN(1x3)" and ``SS-SGN w G-GCN(3x3)" by 0.8\% and 1.0\%, respectively. Learning a global content adaptive graph is more complicated and difficult.



\begin{table}[t]
\small
\centering
\caption{Influence of some techniques on the CS setting of the NTU120 dataset in terms of accuracy (\%) and number of parameters.}
\begin{tabular}{m{0.22\textwidth} p{0.09\textwidth}<{\centering}  p{0.09\textwidth}<{\centering}}
\toprule
Method         & \#Params(M) & CS   \\
\midrule
Baseline       & 0.64     & 67.7   \\
+ DA             & 0.64      & 69.0  \\
+ Velocity       & 0.65      & 75.5  \\
+ Fine-grained movement & 0.66 & 79.1   \\
+ MaxPooling    & 0.66      & 80.9 \\
\bottomrule
\end{tabular}
\label{tab:skill}
\end{table}

\subsubsection{Strong Baseline}
\label{strong}
Previous works usually adopt heavy networks for modeling skeleton sequence of low dimensions \cite{si2018skeleton, si2019attention, shi2019two, zhang2019view}. We exploit some techniques which have been proven very effective in previous works and build a lightweight strong baseline, which has achieved comparable performance as most other state-of-the-art methods \cite{si2018skeleton, zhang2017view, yan2018spatial, gao2019optimized}. We hope this serves as a strong baseline for future research in the skeleton-based action recognition field. All models do not use semantics in this section.

We first build a basic baseline (``Baseline") with the overall pipeline similar to that in Fig.~\ref{fig:framework}. There are three differences. 1) The velocity, fine-grained movement, joint type, and frame index information are not utilized. 2) Data augmentation (DA) (see Data Processing) is not adopted during training. 3) AveragePooling is used instead of Maxpooling as in \cite{yan2018spatial, shi2019two}. 

Table \ref{tab:skill} shows the influence of our adopted techniques for constructing the strong baseline. We have the following three observations. \textbf{1)} Data augmentation improves the performance significantly. Through the augmentation on the observed views, some ``unseen" views could be ``seen" during the training. \textbf{2)} Two stream networks (using both position and velocity) \cite{si2018skeleton} have proven effective, but two separate networks double the number of parameters. We fuse the two types of information in the early stage (in input) and it improves the performance significantly with only a negligible number of additional parameters ({\it i.e.}, 0.01M). \textbf{3)} 
We introduce fine-grained movement to the representation of joints in the same way with velocity. It further improves the performance by 3.6\% with very little cost ({\it i.e.}, additional 0.01M parameters). \textbf{4)} MaxPooling is much more powerful than AveragePooling. The reason is that MaxPooling works like an attention module which drives the network to learn and select discriminative features.

\subsubsection{Effectiveness of Fine-Grained Movement}
\label{sec:fine}

To validate the effectiveness of using the fine-grained movement, we conduct two additional experiments, {\it i.e.}, ``SS-SGN w/o FGM" and ``SS-SGN w CM". ``SS-SGN w/o FGM" denotes that we only use position and velocity information in the DR module. ``SS-SGN w CM" denotes that we replace the fine-grained movement with \emph{coarse movement} by selecting only one referent joint (``spine") of the whole human body. The experimental results are shown in Table~\ref{tab:fine}.

\begin{table}[t]
\small
\centering
\caption{Effectiveness of fine-grained movement on the CS setting of the NTU120 dataset in terms of accuracy (\%) and number of parameters.}
\begin{tabular}{m{0.17\textwidth} p{0.115\textwidth}<{\centering}  p{0.115\textwidth}<{\centering}}
\toprule
Method         & \#Params(M) & CS   \\
\midrule
SS-SGN w/o FGM       & 0.72    & 79.1   \\
SS-SGN w CM            & 0.73      & 79.7  \\
SS-SGN       & 0.73      & 82.3  \\
\bottomrule
\end{tabular}
\label{tab:fine}
\end{table}

\begin{figure*}[!t]
	\begin{center}
		\includegraphics[width=1\linewidth]{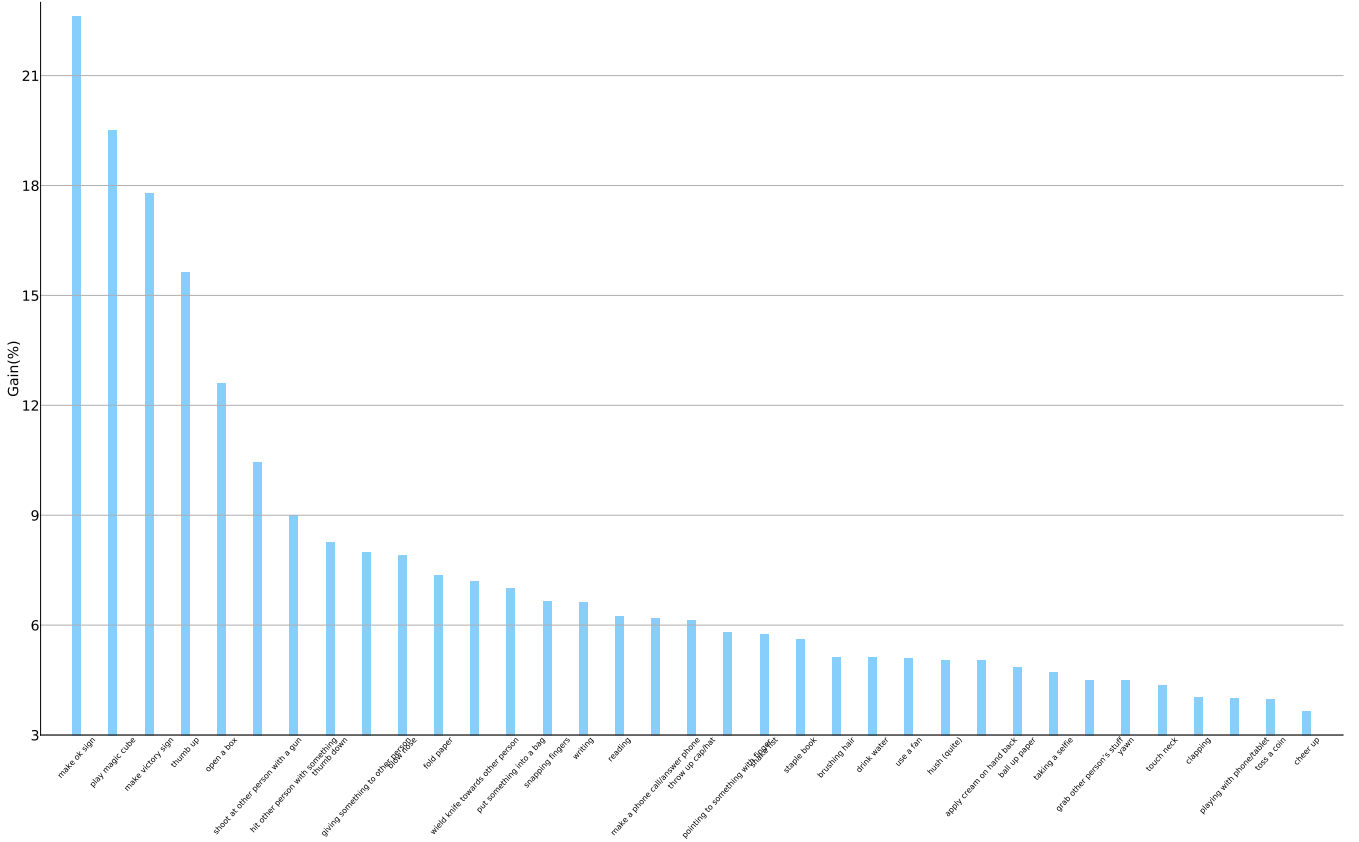}
	\end{center}
	\caption{Effectiveness of using the fine-grained movement on different actions.}
	\label{fig:fine}
\end{figure*}

With the help of fine-grained movement, ``SS-SGN" outperforms ``SS-SGN w/o FGM" by 3.2\% in accuracy. In addition, the coarse movement (``SS-SGN w CM") also brings gain of 0.6\% in comparison with ``SS-SGN w/o FGM", but it is far inferior to fine-grained movement (``SS-SG") by 2.6\%. For the actions with similar poses, the significant difference is the local movement/configuration of some joints, which is much more important than the movement of the joints of the whole human body (``CM").  The fine-grained movement benefits the understanding of the actions with similar skeletons.

To further understand the fine-grained movement, we compare the accuracies of ``SS-SGN" and and ``SS-SGN w/o FGM" and obtain the performance gain for different classes caused by fine-grained movement by subtracting the accuracy of ``SS-SGN" and ``SS-SGN w/o FGM". Fig. \ref{fig:fine} shows the top 55 actions with the largest performance gains.

For \emph{make ok sign} and \emph{make victory sign}, the accuracies of two actions have gains of 22.6\% and 17.8\% with the help of fine-grained movement. For \emph{reading} and \emph{writing}, the accuracies of two actions have gains of 6.2\% and 6.6\%. The fine-grained movement boosts the recognition of those actions with similar skeletons significantly.

\begin{table}[t]
\small
\centering
\caption{Effectiveness of multi-scale strategy on the CS setting of the NTU120 dataset in terms of accuracy (\%) and number of parameters.}
\begin{tabular}{m{0.15\textwidth} p{0.125\textwidth}<{\centering}  p{0.125\textwidth}<{\centering}}
\toprule
Method         & \#Params(M) & CS   \\
\midrule
SS-SGN       & 0.73    & 82.3   \\
MS-SGN(sep.) & 2.19      & 83.4  \\
MS-SGN       & 1.50      & 84.5  \\
\bottomrule
\end{tabular}
\label{tab:multiscale}
\end{table}

\subsubsection{Effectiveness of Mulit-Scale Strategy }
\label{sef:multi}


We design three experiments to validate the effectiveness and efficiency of the proposed multi-scale strategy ``MS-SGN" in Table \ref{tab:multiscale}. ``SS-SGN" denotes there are no multi-scale strategy.  ``MS-SGN~(sep.)" denotes three modules are all not shared while ``MS-SGN" denotes DR module and JL modules are shared while FL modules are not shared across scales. 

We have the following observations.

\noindent 1) ``MS-SGN" outperforms ``SS-SGN" by 2.2\% significantly, which indicates that multi-scale strategy makes the network robust to temporal scale variations. 

\noindent 2) The method of sharing the DR module and JL module and isolating the FL module are more efficient than the method of isolating all the three modules, ``MS-SGN" is superior to ``MS-SGN~(sep.)" by 0.9\% with much fewer parameters. 



\subsubsection{Visualization of SMP}
\label{vis}

\begin{figure}[t]
	\begin{center}
		\includegraphics[width=0.8\linewidth]{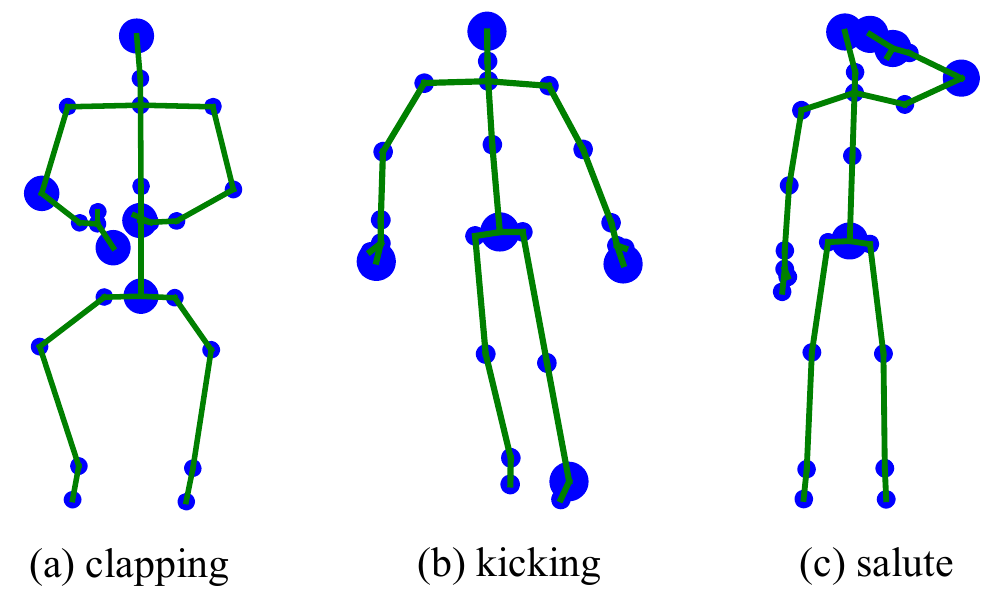}
	\end{center}
	\vspace{-2mm}
	\caption{Visualization of the responses of the spatial MaxPooling layer with respect to three actions, {\it i.e.}, \emph{clapping}, \emph{kicking}, and \emph{salute}.  The top-5 joints selected by SMP are plotted with larger blue circles.}
	\label{fig:vis}
\end{figure}

The spatial Maxpooling (SMP) plays a similar role to the attention mechanism. We visualize the selected joints by SMP for three actions {\it i.e.}, \emph{clapping}, \emph{kicking}, and \emph{salute} in Fig.~\ref{fig:vis}. The dimensions of the responses are 256 and each dimension corresponds to one selected joint. We count the times each joint is selected by the SMP. The top five chosen joints are shown by large blue circles and the rest are shown by small blue circles. We observe that different actions correspond to different informative joints. The left foot is important for \emph{kicking}. Only the left hand is of great value for \emph{salute}, while both left and right hands are essential for \emph{clapping}. These are consistent with human’s perception.

\begin{table}[t] 
   \small
	\centering
	\caption{Performance comparisons on NTU60  with the CS and CV settings in terms of accuracy (\%).}
	\label{tab:ntu}
	\begin{tabular}{m{0.23\textwidth} p{0.05\textwidth}<{\centering}  p{0.05\textwidth}<{\centering} p{0.05\textwidth}<{\centering}}
		\toprule
		{Method}             &Year                             & CS & CV \\
		\midrule
		STA-LSTM \cite{song2017end}                 &2017           & 73.4     & 81.2  \\
		GCA-LSTM \cite{liu2017global}               &2017            & 74.4     & 82.8  \\
		Clips+CNN+MTLN \cite{ke2017new}             &2017          & 79.6     & 84.8  \\
		VA-LSTM \cite{zhang2017view}                &2017            & 79.4     & 87.6  \\
		ElAtt-GRU\cite{zhang2018adding}            &2018            & 80.7     & 88.4 \\
		ST-GCN \cite{yan2018spatial}                &2018             & 81.5     & 88.3 \\
		DPRL+GCNN \cite{tang2018deep}               &2018            & 83.5     & 89.8 \\
		SR-TSL  \cite{si2018skeleton}               &2018           & 84.8     & 92.4 \\
		HCN \cite{li2018co}                         &2018           & 86.5     & 91.1 \\
		AGC-LSTM (joint) \cite{si2019attention}      &2019           & 87.5     & 93.5 \\
		AS-GCN \cite{li2019actional}                &2019           & 86.8     & 94.2 \\
		GR-GCN     \cite{gao2019optimized}           &2019          & 87.5     & 94.3 \\          
		VA-CNN  \cite{zhang2019view}                &2019           & 88.7     & 94.3 \\
	    SGN \cite{zhang2020semantics}               &2020           & 89.0      &94.5 \\
		MS-G3D(Joint) \cite{liu2020disentangling}           &2020           & 89.4     & 95.0 \\
		1s Shift-GCN \cite{cheng2020skeleton}     &2020           &87.8    &95.1 \\
 		\midrule
		SS-SGN               &-        & 89.6    &94.6  \\
        MS-SGN             &-   & \textbf{90.1}     & \textbf{95.2} \\
		\bottomrule
	\end{tabular}
	\vspace{3mm}
\end{table}

\subsubsection{Complexity of SGN}
\label{efficiency}

We discuss the complexity of SGN by comparing it with eight state-of-the-art methods for skeleton-based action recognition. As shown in Fig.~\ref{fig:paras}, the number of parameters of VA-RNN \cite{zhang2019view} is the least, but the accuracy is the poorest. VA-CNN\cite{zhang2019view} and 2s-AGCN\cite{shi2019two} achieve good accuracy, but the numbers of parameters are so large. In comparison with the RNN-based, GCN-based, and CNN-based methods, our proposed SGN achieves slightly better performance with much fewer parameters, which makes SGN attractive for many practical applications which have limited computational power.

\subsection{Comparison with the State-of-the-arts}
\label{compare}

We compare the proposed ``SS-SGN" and ``MS-SGN" with other state-of-the-art methods on the NTU60, NTU 120, and SYSU datasets in Table \ref{tab:ntu}, Table \ref{tab:ntu120}, and Table \ref{tab:SYSU}, respectively.

\setlength{\tabcolsep}{5pt}
\begin{table}[t] 
    \small
	\centering
	\caption{Performance comparisons on  NTU120 with the C-Subject and C-Setup settings in terms of accuracy (\%).}
	\begin{tabular}{lcccc}
		\toprule
		{Method}             &Year                             & C-Subject & C-Setup \\
		\midrule
		Part-Aware LSTM   \cite{shahroudy2016ntu}   &2016            & 25.5    & 26.3  \\
		ST-LSTM + Trust Gate \cite{liu2016spatio}   &2016            & 55.7    & 57.9  \\
		GCA-LSTM \cite{liu2017global}               &2017            & 58.3    & 59.2  \\
		Clips+CNN+MTLN \cite{ke2017new}             &2017            & 58.4    & 57.9  \\
    	Two-Stream GCA-LSTM \cite{liu2017skeleton} &2017             &61.2     &63.3 \\
		RotClips+MTCNN \cite{ke2018learning}        &2018            &62.2     &61.8 \\
		Body Pose Evolution Map \cite{liu2018recognizing} &2018      &64.6     &66.9 \\
 		2s-AGCN \cite{shi2019two}    &2019                           &82.9     &84.9 \\
 		SGN \cite{zhang2020semantics} & 2020  & 79.2   & 81.5 \\
 		1s Shift-GCN  \cite{cheng2020skeleton} & 2020                &80.9     &83.2 \\
		\midrule
		SS-SGN    &-   &82.3            &83.4 \\
        MS-SGN             &-   &\textbf{84.5}   & \textbf{85.6} \\
		\bottomrule
	\end{tabular}
	\label{tab:ntu120}
	\vspace{2mm}
\end{table}

\begin{table}[t]
   \small
	\centering
	\caption{Performance comparisons on SYSU  in terms of accuracy (\%). * denotes the model uses parameters pre-trained on NTU60.}
		\begin{tabular}{m{0.23\textwidth} p{0.05\textwidth}<{\centering}  p{0.05\textwidth}<{\centering} p{0.05\textwidth}<{\centering}}
		\toprule
		Method  & Year & CS  & SS \\
		\midrule
		VA-LSTM \cite{zhang2017view}       &2017          & 77.5  & 76.9 \\
		ST-LSTM \cite{liu2018skeletontrust}  &2018      & 76.5 & - \\
		GR-GCN     \cite{gao2019optimized}    &2019      & 77.9 & - \\
		Two stream GCA-LSTM  \cite{liu2017skeleton}       &2017       & 78.6 & - \\
		SR-TSL  \cite{si2018skeleton}         &2018      &81.9  & 80.7 \\
		ElAtt-GRU* \cite{zhang2018adding}      &2018          & 85.7 & 85.7 \\
		SGN* \cite{zhang2020semantics}  &2020    & 90.6  &89.3 \\
		\midrule
        SS-SGN     & - & 85.1 & 83.0 \\
        MS-SGN     & - & 84.3 & 82.8 \\
        SS-SGN*    & -  & 91.9 & 90.2 \\
        MS-SGN*    & -  & \textbf{92.8} & \textbf{91.3} \\
		\bottomrule      
		\label{tab:SYSU}
	\end{tabular}
\end{table}

As shown in Table \ref{tab:ntu}, ``AGC-LSTM(joint)" \cite{si2019attention} and ``VA-CNN" \cite{zhang2019view} are two representative methods for RNN-based and CNN-based methods, respectively. ``MS-SGN" outperforms them by 2.6\% and 0.4\% in accuracy for the CS setting but  uses only ten percent of their numbers of parameters as shown in Fig.~\ref{fig:paras}. When compared to the GCN-based methods \cite{shi2019two,liu2020disentangling, cheng2020skeleton}, the  proposed ``MS-SGN" achieves competitive performance and outperforms them in the number of parameters, training and inference speed.


As shown in Table \ref{tab:ntu120} and Table \ref{tab:SYSU}, the proposed SGN achieves the best accuracy on NTU120 and SYSU. It should be noted that \cite{zhang2018adding} and \cite{zhang2020semantics} used the pre-trained model on the NTU60 dataset to initialize parameters for the SYSU dataset. The proposed ``SS-SGN" and ``MS-SGN" achieves the best performance with or without the pre-trained model.

\section{Conclusions}

In this work, we have presented a simple yet effective end-to-end multi-scale semantics-guided neural network for high performance skeleton-based human recognition. We explicitly introduce the high level semantics, {\it i.e.}, joint type and frame index, as part of the network input. To model the correlations of joints, we have proposed a joint-level module for capturing the correlations of joints in the same frame and a frame-level module for modeling the dependencies of frames where all joints in the same frame are taken as a whole. Semantics helps improve the capability of both the GCN and CNN. To be robust to the temporal scale variations, we propose a multi-scale strategy by using multi-scale frame-level modules. In addition, we have developed a strong baseline which is better than most previous methods. The fine-grained movement of joints, which is incorporated in the strong baseline,  facilitates the understanding of fine-grained actions.  With an order of magnitude smaller model size than some previous works, our proposed model achieves the state-of-the-art results on three benchmark datasets.

\section*{Acknowledgment}
This work was partially supported by the National Natural Science Foundation of China (Grant No. 61751308 and 61773311).


%

%



\bibliographystyle{abbrv}
\bibliography{egbib}
\ifCLASSOPTIONcaptionsoff
  \newpage
\fi

\end{document}